\documentclass{article}

\usepackage{microtype}
\usepackage{graphicx}
\usepackage{subcaption}
\usepackage{booktabs}
\usepackage{hyperref}

\usepackage[accepted]{icml2026}
\usepackage[T1]{fontenc}
\usepackage{amsmath}
\usepackage{amssymb}
\usepackage{mathtools}
\usepackage{amsthm}
\usepackage[capitalize,noabbrev]{cleveref}

\theoremstyle{plain}

\theoremstyle{definition}

\theoremstyle{remark}

\icmltitlerunning{CASCADE Conformal Prediction}

\begin{document}

\twocolumn[
  \icmltitle{CASCADE Conformal Prediction: Uncertainty-Adaptive Prediction Intervals for Two-Stage Clinical Decision Support}

  \icmlsetsymbol{equal}{*}

  \begin{icmlauthorlist}
    \icmlauthor{Ricardo Diaz-Rincon}{ufl}
    \icmlauthor{Muxuan Liang}{mda}
    \icmlauthor{Adolfo Ramirez-Zamora}{lou}
    \icmlauthor{Benjamin Shickel}{ufl}
  \end{icmlauthorlist}

  \icmlaffiliation{ufl}{University of Florida}
  \icmlaffiliation{mda}{MD Anderson Cancer Center}
  \icmlaffiliation{lou}{University of Louisville}

  \icmlcorrespondingauthor{Benjamin Shickel}{shickelb@ufl.edu}

  \icmlkeywords{conformal prediction, uncertainty quantification, two-stage systems, Parkinson's disease, clinical decision support}

  \vskip 0.1in
]

\printAffiliationsAndNotice{}

\begin{abstract}
Effective medication management in Parkinson's Disease (PD) is challenging due to
heterogeneous disease progression, variable patient response, and medication side
effects. While AI models can forecast levodopa equivalent daily dose (LEDD) as a
measure of medication needs, standard uncertainty quantification often fails to
communicate the reliability of these predictions, treating high and low confidence
clinical decisions identically. We introduce CASCADE (Calibrated Adaptive Scaling
via Conformal And Distributional Estimation), a novel conformal prediction framework
that propagates epistemic uncertainty from a screening classifier to adapt downstream
predictions. Unlike standard conformal methods that rely on auxiliary residual
regression, we leverage epistemic uncertainty from a primary classification task
(identifying whether a medication change is needed) to dynamically scale the
prediction intervals of a secondary regression task (predicting how much change).
By mapping Venn-Abers multi-probabilistic uncertainty directly to non-conformity
scores, our framework achieves continuous risk adaptation. We demonstrate that this
``cascade effect'' produces highly efficient intervals for confident patients (38.9\%
narrower than standard conformal baselines) while automatically expanding intervals
to ensure robust coverage for uncertain cases, bridging the gap between discrete
clinical decision-making and continuous dose forecasting in PD. 
\end{abstract}

\section{Introduction}
\label{sec:intro}

\begin{figure*}[t!]
    \centering
    \includegraphics[width=\textwidth]{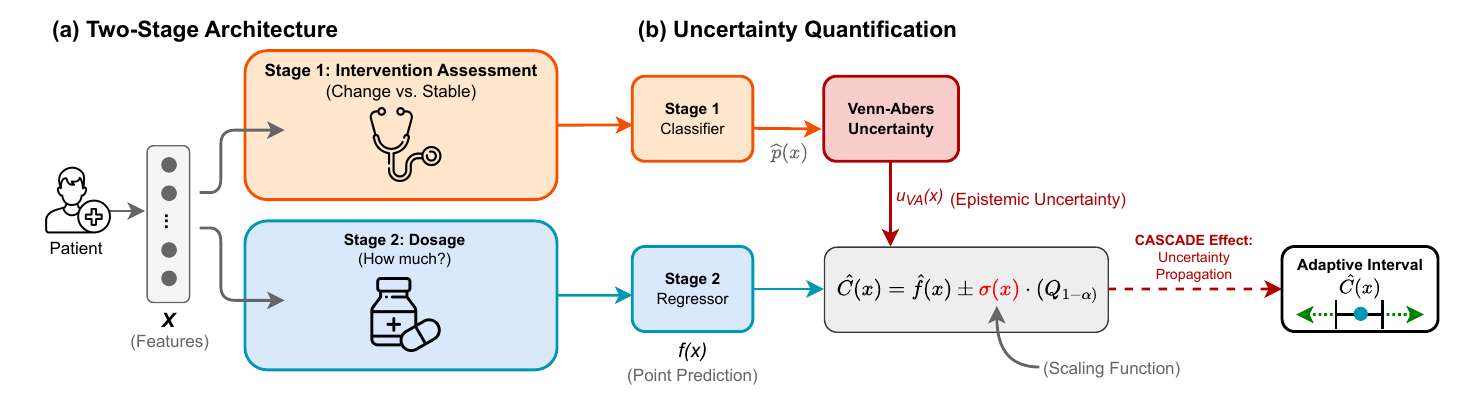}
    \caption{Overview of the CASCADE Framework.
    \textbf{(a) Two-Stage Architecture:} A Stage 1 classifier identifies patients
    requiring medication adjustment, followed by a Stage 2 regressor that predicts
    the dosage.
    \textbf{(b) Uncertainty Quantification:} CASCADE extracts the epistemic uncertainty
    score $u_{\text{VA}}(x)$ directly from the Stage 1 predicted probability $\hat{p}(x)$.
    This score flows into the scaling function $\sigma(x)$ (highlighted in red), which
    scales the prediction interval. This mathematical propagation drives the ``cascade
    effect'' (red dashed arrow), where the final adaptive interval $\hat{C}(x)$ expands
    or shrinks based on patient uncertainty.}
    \label{fig:cascade_overview}
\end{figure*}

Parkinson's Disease (PD) medication management is a high-stakes sequential decision
problem: first determining whether a dosage adjustment is needed, then quantifying
how much. Dosing errors carry direct patient risk: inadequate titration leaves
patients with debilitating motor symptoms, while over-medication can trigger
levodopa-induced dyskinesia (LID) and neuropsychiatric complications
\citep{Armstrong2020, Espay2016}. The levodopa equivalent daily dose (LEDD)
standardizes medication burden across drug classes \citep{Tomlinson2010}, but optimal
titration remains a subjective, trial-and-error process heavily reliant on brief
clinical assessments \citep{Armstrong2020}.

Recent advancements in artificial intelligence (AI) have accelerated the adoption of
clinical decision support systems. Because most clinical decisions follow a hierarchical
logic: first determining \textit{if} an intervention is needed before deciding
\textit{how much}, these systems increasingly rely on two-stage architectures. In PD,
such a system operates in two stages: As shown in Figure~\ref{fig:cascade_overview}(a), first, a classifier identifies if a patient’s medication regimen requires adjustment; second, a regression model predicts how much
the dosage should change \citep{diaz-rincon25b}. However, applying uncertainty
quantification to these two-stage architectures presents a critical challenge:
information loss at the decision boundary. While conformal prediction (CP) provides
distribution-free coverage guarantees for such models \citep{vovk2005, Shafer2008},
it typically treats the regression stage in isolation, regardless of first-stage
uncertainty.

\begin{figure*}[t!]
    \centering
    \includegraphics[width=\textwidth]{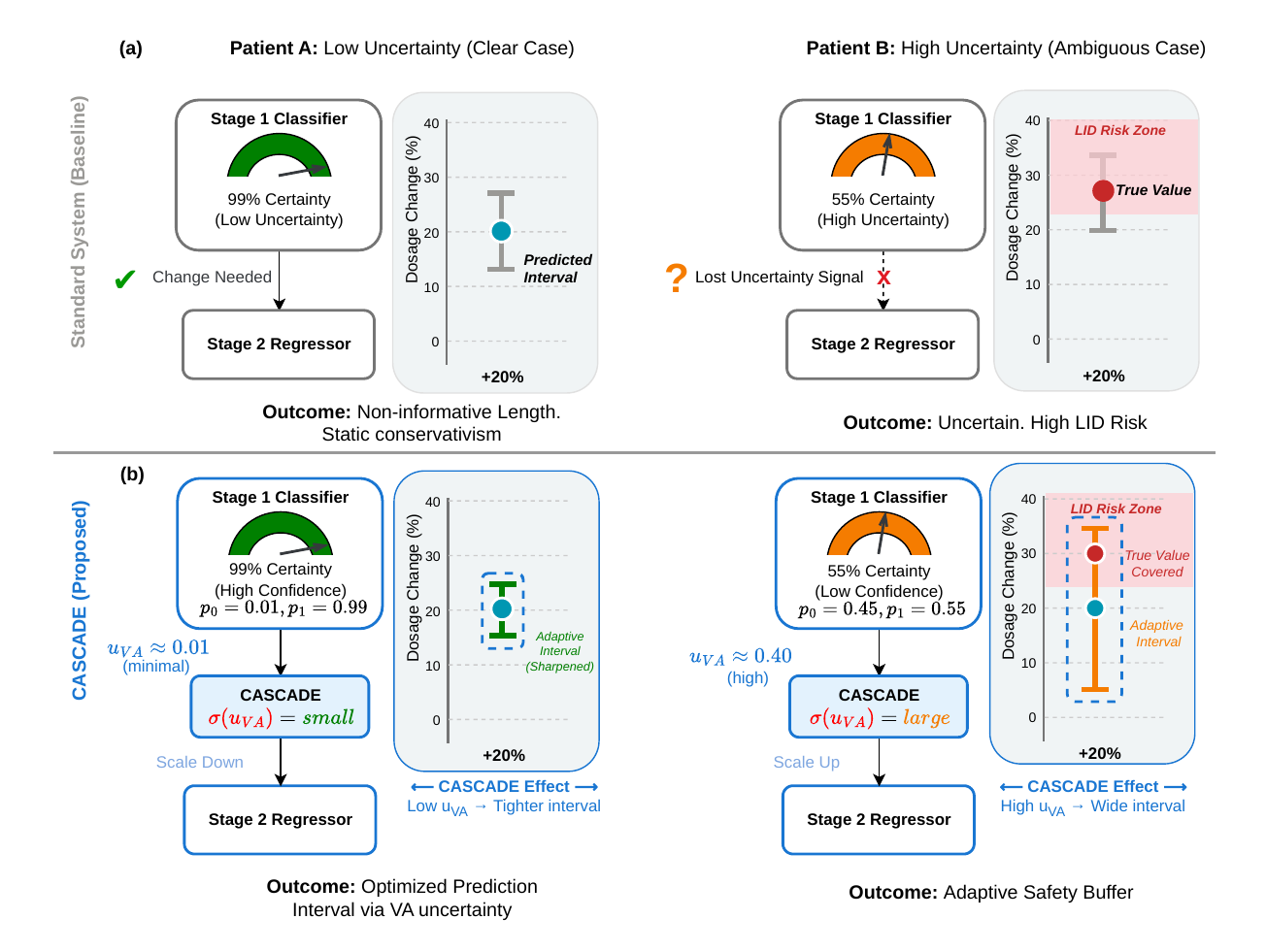}
    \caption{The CASCADE Mechanism.
    \textbf{(a) Standard System:} The regression model is blind to upstream confidence,
    producing non-informative, fixed-length prediction intervals for both the low
    uncertainty Patient A ($p=0.99$) and the high uncertainty (ambiguous) Patient B
    ($p=0.55$). While actionable for A, this rigidity is dangerous for B: the model
    confidently predicts a dosage increase despite high ambiguity. This erroneous
    precision carries a high risk of over-medication, potentially driving the patient
    into the levodopa-induced dyskinesia ``(LID) Risk Zone''.
    \textbf{(b) CASCADE (Proposed):} The framework solves this by explicitly coupling
    the two stages, translating the epistemic uncertainty from Stage 1 into Stage 2.
    Left (Patient A): High confidence results in a minimal Venn-Abers score ($u_{\text{VA}}$).
    The mechanism responds with compressive scaling, producing an optimized prediction,
    eliminating unnecessary conservatism and confirming the dosage target is safe.
    Right (Patient B): The increased uncertainty yields a high ($u_{\text{VA}}$). The
    mechanism responds with expansive scaling, creating an adaptive safety buffer that
    covers the ground truth and explicitly signals that a standard dosage increase
    carries significant risk.}
    \label{fig:mechanism}
\end{figure*}

Consider a neurologist using such a system for two patients: For patient A, the system
is 99\% certain a medication change is needed. For patient B, the system is ambivalent,
predicting a change with 55\% certainty (hovering near the 0.5 decision boundary). A
standard regression model, blind to this upstream wavering, treats these two patients
identically, resulting in static conservatism. It might predict a 20\% medication
increase for both, assigning similar confidence intervals. However, the clinical risk
profile is vastly different: the prediction for patient A is actionable, while the
prediction for patient B is uncertain, carrying a higher risk of errors that could
potentially lead to levodopa-induced dyskinesia \citep{Espay2016, Olanow2006,
Jenner2008} and other complications \citep{newman2009parkinsonism,
taylor2016neuropsychiatric}. This scenario, illustrated in Figure~\ref{fig:mechanism}(a), exposes a critical gap
that limits the adoption of uncertainty quantification in high-stakes healthcare settings.

To bridge this gap, we introduce CASCADE (Calibrated Adaptive Scaling via Conformal
And Distributional Estimation), a framework that explicitly couples classification
uncertainty with regression calibration. Instead of treating these stages independently,
we propose that the epistemic uncertainty of the Stage 1 decision, measured through
Venn-Abers (VA) predictors \citep{Vovk2012}, should serve as a dynamic scaling factor
for the Stage 2 prediction intervals. This creates a ``cascade effect'': as the
system's uncertainty increases, the prediction intervals automatically widen to reflect
this. Clinically, this ensures that the model's reliability scales with the difficulty
of the case, acting as a safeguard by preventing dangerous overconfidence in complex
scenarios while maintaining high precision for straightforward patients.

\textbf{Our contributions.} (1) We introduce the CASCADE framework, creating a
mechanism to map Venn-Abers uncertainty directly to conformal non-conformity scores.
This enables cross-task uncertainty transfer, leveraging classification reliability to
scale regression intervals without the computational overhead of training auxiliary
error models. (2) We implement CASCADE using two distinct strategies: a discrete
Mondrian approach and a Continuous Cascade approach using mean-centered scaling.
Within the continuous implementation we introduce a tunable scaling parameter
(${\beta}$), allowing practitioners to explicitly calibrate the system's sensitivity
to uncertainty. We demonstrate that this continuous parameterization removes the
discretization artifacts and sample-size fragmentation inherent in traditional binning
approaches. (3) We validate this approach on a decade of PD inpatient data from 631 patients at the University of Florida (UF) Health, showing
that CASCADE maintains marginal coverage guarantees while reducing interval length by
38.9\% for high-confidence patients compared to baselines.

\section{Related Work}
\label{sec:related_work}

\paragraph{Machine Learning in PD Medication Management.}
To facilitate medication decisions, recent approaches have applied machine learning
to forecast medication needs through the LEDD \citep{diaz-rincon24}. These range from
predicting raw levodopa doses \citep{Salmanpour2023} and post-surgical adjustments
\citep{Shamir2015} to clustering medication regimes \citep{Watts2021} and optimizing
combinations via reinforcement learning \citep{Kim2021}. While these approaches achieve
adequate accuracy, they predominantly rely on point predictions without accounting for
prediction uncertainty. In high-stakes clinical settings, a single deterministic value
without reliability bounds is insufficient, as it fails to distinguish between a robust
forecast and a statistical guess. Without uncertainty bounds, models may encourage
overconfident interventions in ambiguous cases, potentially leading to suboptimal
patient outcomes \citep{Beli2019}. Consequently, there is a critical need for rigorous
uncertainty quantification mechanisms that can translate model confidence into
clinically interpretable safety bounds.

\paragraph{Conformal and Adaptive Conformal Prediction.}
Conformal prediction \citep{vovk2005, Shafer2008} addresses this need by providing
prediction intervals with distribution-free, finite-sample coverage guarantees.
Methods include Na\"{i}ve, Standard (Split), Cross-Validation (CV+) \citep{Barber2021}
and Jackknife+-After-Bootstrap (J+aB) \citep{Kim2020}. However, marginal validity
can mask conditional failure: a model may achieve 80\% marginal coverage by
over-covering ``easy'' patients (samples) while systematically under-covering
``difficult'' ones. We implement these four baseline approaches for comparison.
Comprehensive details about these methods are provided in Appendix~\ref{sec:cp_framework}.
To address heterogeneity, Mondrian (or Group-Balanced) CP \citep{vovk2005} achieves
group-conditional validity by partitioning data into strata (e.g., by disease stage)
and calibrating intervals independently within each group. Complementarily, Normalized
(or locally scaled) CP \citep{Lei2018, Papadopoulos2011} focuses on adaptivity, scaling
interval length based on a difficulty estimate $\sigma(x)$. However, both strategies
primarily focus on single-stage prediction tasks without addressing the challenge of
uncertainty propagation in cascaded systems, where an upstream classification decision
inherently limits the reliability of downstream predictions.

\paragraph{Uncertainty Quantification in Two-Stage Systems.}
In clinical settings, prediction often follows a two-stage logic: first determining if
an intervention is needed (classification), then assessing how much (regression). While
multi-stage architectures effectively handle zero-inflated data by separating decision
and prediction tasks \citep{diaz-rincon25a}, they often introduce information loss at
the decision boundary. Current frameworks typically decouple uncertainty quantification
across these stages, utilizing the classification threshold as a hard boundary: once a
patient crosses the decision threshold, the probabilistic ambiguity of that decision is
discarded.

Consequently, the epistemic uncertainty of the Stage 1 decision (``Does this patient
truly require a medication change?'') is rarely propagated to Stage 2 (``How much
change?''). This creates a reliability mismatch, where predictions for borderline cases
appear deceptively confident, failing to reflect the compounded risk of the hierarchical
decision process. Our work bridges this gap by introducing a cascade mechanism that
prevents this information loss, directly mapping classification reliability to downstream
regression.

\section{Methods}
\label{sec:methods}

\subsection{Problem Formulation}

Consider a two-stage prediction setting characterized by a dataset
$\mathcal{D} = \{(x_i, y_i)\}_{i=1}^{n}$. Where $x_i \in \mathbb{R}^d$ is a feature
vector encoding patient characteristics (e.g., age, demographics, clinical variables),
and the target $y_i$ is governed by a latent decision process followed by a value
estimation. While the framework applies to any such system, we demonstrate it on
Parkinson's Disease medication management, where $y_i \in \mathbb{R}$ denotes the
percentage change in LEDD ($y_i = 0$ indicates no medication adjustment). To perform
conformal calibration, we randomly partition the data into a proper training set
$\mathcal{D}_{\text{train}}$ and a hold-out calibration set $\mathcal{D}_{\text{cal}}$
(comprising 20\% of $\mathcal{D}$). Let $\hat{f}$ denote the underlying regression
model trained on $\mathcal{D}_{\text{train}}$, which outputs a prediction $\hat{f}(x)$
for a given input.

Our objective is to construct a prediction set $\hat{C}(x_{\text{new}})$ that satisfies
marginal coverage:
\begin{equation}
  \mathbb{P}(y_{\text{new}} \in \hat{C}(x_{\text{new}})) \geq 1 - \alpha
  \label{eq:coverage}
\end{equation}
where $1 - \alpha$ is the confidence level (e.g., 80\%). Crucially, beyond this
marginal guarantee, we seek to improve local adaptivity, ensuring the interval length
dynamically scales with the reliability of the upstream classification decision.

\subsection{The CASCADE Framework}

To address the information loss in standard two-stage systems, we introduce CASCADE
(Calibrated Adaptive Scaling via Conformal And Distributional Estimation). The core
principle of this framework is cross-task uncertainty transfer: we postulate that the
epistemic uncertainty of Stage 1, which identifies whether a patient requires medication
changes, is a critical signal for the reliability of the downstream regression task
(Stage 2), which predicts the specific dosage.

Instead of treating the stages as statistically independent, CASCADE propagates the
reliability of the classification step directly into the conformal calibration process. As illustrated in Figure~\ref{fig:cascade_overview}(b), the framework operates in two distinct phases: (1)~\textbf{Epistemic Uncertainty Quantification (Stage 1):}
quantifying the ambiguity of the binary classifier using Venn-Abers calibration; and
(2)~\textbf{Adaptive Conformal Calibration (Stage 2):} scaling the regression
prediction intervals based on the Stage 1 uncertainty.

\subsection{Stage 1: Epistemic Uncertainty Estimation}

We first train a classifier to estimate the predicted probability of the event of
interest (e.g., medication change), denoted as $\hat{p}(x)$. To quantify the epistemic
uncertainty of this decision, we employ Venn-Abers predictors \citep{Vovk2012}. Unlike
standard probability estimates (e.g., softmax) which produce a single scalar and are
often poorly calibrated in non-linear models \citep{Guo2017}, Venn-Abers provides a
rigorous multi-probabilistic interval $([p_0(x), p_1(x)])$. This interval explicitly
captures epistemic uncertainty without requiring distributional assumptions.

We define the Venn-Abers Uncertainty Score $u_{\text{VA}}(x)$ as the length of this
probability interval:
\begin{equation}
  u_{\text{VA}}(x) = p_1(x) - p_0(x).
  \label{eq:uva}
\end{equation}
$u_{\text{VA}}(x)$ serves as a rigorous proxy for the stability of the clinical
decision: a wide interval implies high ambiguity in whether the patient truly requires
adjustment, regardless of the predicted class probability.

\subsection{Stage 2: Adaptive Conformal Calibration}

For the regression task (Stage 2), we focus on patients where medication adjustments
are indicated. In our experiments, to isolate the performance of the conformal
calibration from upstream classification errors, we perform our evaluation on the
subset of patients requiring medication changes as indicated by their ground truth
class ($y_i \neq 0$). Additional experiments for $\hat{y} \neq 0$ (using the predicted
probability of medication changes) can be found in Appendix~\ref{sec:predicted_patients}.

We propose two distinct strategies for constructing uncertainty-aware intervals:

\subsubsection{Mondrian (Discrete Stratified CP)}

This implementation utilizes Mondrian Conformal Prediction \citep{vovk2005} as our
discrete baseline. We a priori partition the calibration set $\mathcal{D}_{\text{cal}}$
into $K$ disjoint strata $G_1, \ldots, G_K$ based on fixed quantiles of the uncertainty
score $u_{\text{VA}}(x)$. In our experiments, we evaluate granularities of $K = 3$
(tertiles). Additional results for $K \in \{5, 7\}$ can be found in
Appendix~\ref{sec:granularity} Table~\ref{tab:mondrian_sensitivity}. This method
creates independent calibration pools, ensuring validity within each stratum although
suffering from data fragmentation (reduced effective sample size).

For a new patient $x_{\text{new}}$, we identify the stratum $k$ such that
$u_{\text{VA}}(x_{\text{new}}) \in G_k$. The prediction interval is constructed using
only the residuals of calibration examples in that same uncertainty stratum:
\begin{equation}
  \hat{C}_{\text{strat}}(x_{\text{new}}) = \left[\hat{f}(x_{\text{new}}) - q_r^{(k)},
  \;\hat{f}(x_{\text{new}}) + q_r^{(k)}\right],
  \label{eq:mondrian}
\end{equation}
where $q_r^{(k)}$ is the $(1-\alpha)$ quantile of the absolute residuals
$|y_i - \hat{f}(x_i)|$ computed solely from the sub-population in $G_k$. While valid,
this method reduces the effective sample size for calibration to $N_{\text{cal}}/K$,
a known limitation of Mondrian CP that can increase variance in the estimated quantiles
\citep{Angelopoulos2021}.

\subsubsection{Continuous CASCADE (Mean-Centered Scaling)}

To avoid discretization artifacts and sample fragmentation, we propose a continuous
scaling mechanism based on Normalized (Locally Scaled) Conformal Prediction
\citep{Lei2018}. We define a scaling function $\sigma(x)$ that adapts the interval
length relative to the population's average uncertainty, parameterized by a sensitivity
factor $\beta \geq 0$:
\begin{equation}
  \sigma(x) = 1 + \beta \left(\frac{u_{\text{VA}}(x)}{\bar{u}_{\text{VA}}} - 1\right),
  \label{eq:sigma}
\end{equation}
where $\bar{u}_{\text{VA}}$ is the mean VA uncertainty of the calibration set. This
mean-centered scaling creates a logical pivot: when $u_{\text{VA}}(x) \approx
\bar{u}_{\text{VA}}$, $\sigma(x) \approx 1$ (standard length); when
$u_{\text{VA}}(x) > \bar{u}_{\text{VA}}$, $\sigma(x) > 1$ (interval expands); and when
$u_{\text{VA}}(x) < \bar{u}_{\text{VA}}$, $\sigma(x) < 1$ (interval shrinks).
We define the scaled non-conformity scores for each calibration point $i$ as:
\begin{equation}
  S_i = \frac{|y_i - \hat{f}(x_i)|}{\sigma(x_i)}.
  \label{eq:scores}
\end{equation}
We then compute $Q_{1-\alpha}$, which is the $(1-\alpha)$-th empirical quantile of the
normalized scores $\{S_i\}$ across the full calibration set. The final prediction
interval is:
\begin{equation}
\begin{split}
  \hat{C}_{\text{cont}}(x_{\text{new}}) = \Bigl[
    &\hat{f}(x_{\text{new}}) - Q_{1-\alpha} \cdot \sigma(x_{\text{new}}),\\
    &\hat{f}(x_{\text{new}}) + Q_{1-\alpha} \cdot \sigma(x_{\text{new}})\Bigr].
\end{split}
\label{eq:cascade_interval}
\end{equation}
Continuous CASCADE uses the entire calibration set ($n$ samples) to estimate the single
non-conformity quantile $Q_{1-\alpha}$, whereas Mondrian CP splits the data into small
bins. This global estimation ensures stable scaling factors and maintains robust
statistical power, allowing the interval length to scale continuously with the
difficulty of the clinical decision, as illustrated in Figure~\ref{fig:mechanism}(b).

\section{Evaluation Metrics}
\label{sec:metrics}

\subsection{Conformal Metrics}

We assess model performance on the test set filtered on $y_i \neq 0$ (i.e., patients
requiring medication changes as indicated by their ground truth class). This focuses
the evaluation on actionable cases, measuring how well the prediction intervals cover
the ground truth dosage change given that an intervention is necessary. We employ
three metrics:

\textbf{Marginal Coverage} ($1 - \alpha$): Defined as the proportion of test samples
where the true outcome lies within the predicted interval. We target a nominal coverage
level 80\% ($\alpha = 0.2$). This value aligns with standard benchmarks in conformal
prediction literature \citep{Angelopoulos2023, vovk2005} and prioritizes the
reliability-precision equilibrium required for clinical utility \citep{Svensson2018,
Papadopoulos2017}. Ablation studies supporting our choice of $\alpha$ are in
Appendix~\ref{sec:alpha_ablation}.

\textbf{Average Length (Efficiency):} The mean size of the prediction interval.
Shorter intervals are preferred, provided they satisfy coverage guarantees.

\textbf{Cascade Ratio (Adaptivity):} To quantify the framework's ability to
differentiate between ``safe'' and ``ambiguous'' cases, we introduce the Cascade Ratio
(CR). To ensure a consistent comparison across experiments with varying granularity
$K$, we define CR as the ratio of the average interval length in the highest uncertainty
partition (top $1/K$-th) to that of the lowest uncertainty partition (bottom $1/K$-th).
For instance, $K = 3$ compares the top 33\% against the bottom 33\%. For continuous
CASCADE, we evaluate the metric on the subset of patients falling into these same
uncertainty quantiles to maintain comparability with the discrete baselines.
\begin{equation}
  \text{CR} = \frac{\mathbb{E}[\text{len}(\hat{C}(x)) \mid
    u_{\text{VA}}(x) \geq Q_{1-1/K}]}
  {\mathbb{E}[\text{len}(\hat{C}(x)) \mid u_{\text{VA}}(x) \leq Q_{1/K}]},
  \label{eq:cr}
\end{equation}
where $Q_\tau$ denotes the $\tau$-th empirical quantile of the uncertainty scores in
the test set. A higher CR indicates superior conditional adaptivity, implying the model
correctly expands intervals for difficult decisions while narrowing them for clear ones.

\subsection{Statistical Validation}

To verify the observed adaptivity represents a genuine improvement rather than random
variance, we perform a suite of normative statistical tests:

\textbf{Kolmogorov-Smirnov (KS) Test:} We utilize it to quantify the distributional
divergence between the interval lengths of CASCADE and the Standard CP baseline. A
significant statistic ($D$) confirms that our scaling mechanism successfully alters the
interval structure, producing a statistically distinct distribution of interval lengths
driven by uncertainty rather than the static profile of the baseline.

\textbf{Spearman Correlation:} To confirm that the interval length is driven by the
uncertainty scores, we evaluate the rank correlation ($\rho$) between the interval
length and the Venn-Abers score ($u_{\text{VA}}$). A strong positive correlation
validates that prediction intervals scale with model uncertainty.

\textbf{Winkler Score:} Computed to evaluate the joint trade-off between coverage and
interval length (lower is better). This proper scoring rule penalizes wide intervals
and applies a penalty for non-coverage (true value is outside the interval), providing
a single summary statistic for model utility that balances safety with efficiency.

\section{Results}
\label{sec:results}

\subsection{Experimental Setup}

For both Stage 1 classification and Stage 2 regression we employ XGBoost models
\citep{Chen2016} optimized via Bayesian Optimization \citep{Wu2019}. Code is available at \url{https://github.com/rdiazrincon/cascade_conformal_pd}. Detailed
hyperparameters and training configurations are provided in Appendix~\ref{sec:model}.
We evaluate on a cohort of 631 inpatient admissions from UF Health. Further details about patient cohort and medications can be found in Appendix~\ref{sec:cohort}.
Additional experiments with different models (Logistic/Linear Regression) can be found
in Appendix~\ref{sec:alt_models}.

\subsection{Overall Performance Comparison}

We evaluate the performance of CASCADE against four conformal benchmarks (Na\"{i}ve,
Standard CP, CV+, J+aB) and the discrete Mondrian baseline. Table~\ref{tab:overall}
summarizes the comparative performance.

\begin{table*}[t]
  \caption{Overall Performance Comparison. Continuous CASCADE achieves the optimal
  balance between coverage and adaptivity (CR 4.23). Standard baselines produce
  fixed-length intervals, resulting in no conditional adaptivity (CR 1.00).}
  \label{tab:overall}
  \begin{center}
    \begin{footnotesize}
      \begin{tabular}{lccc}
        \toprule
        Method & Marg.\ Cov.$^\ddagger$ & Length$^\ddagger$ & CR$^{\dagger\ddagger}$ \\
        \midrule
        \multicolumn{4}{l}{\textit{Standard Baselines}} \\
        Na\"{i}ve   & 52.5\% [0.469--0.583] & 0.031 [0.031--0.031] & 1.00 [1.00--1.00] \\
        Standard CP & 84.0\% [0.801--0.879] & 0.113 [0.113--0.113] & 1.00 [1.00--1.00] \\
        CV+         & 83.5\% [0.792--0.876] & 0.100 [0.096--0.105] & 1.06 [0.94--1.20] \\
        J+aB        & 60.6\% [0.547--0.658] & 0.132 [0.128--0.137] & 0.97 [0.89--1.04] \\
        \midrule
        \multicolumn{4}{l}{\textit{CASCADE Framework}} \\
        Mondrian ($K=3$)  & 86.5\% [0.824--0.902] & 0.118 [0.113--0.123] & 2.02 [2.02--2.02] \\
        \textbf{Cont.\ CASCADE ($\beta=0.7$)} & \textbf{80.1\%} [0.756--0.844] & \textbf{0.148} [0.135--0.162] & \textbf{4.23} [3.82--4.70] \\
        \bottomrule
      \end{tabular}
    \end{footnotesize}
  \end{center}
  \vskip -0.1in
  {\footnotesize $^\dagger$CR: avg length (Top 33\% Unc.) / (Bottom 33\% Unc.) \quad
  $^\ddagger$95\% bootstrap confidence intervals}
\end{table*}

\textbf{Baseline Selection.} As shown in Table~\ref{tab:overall}, both Na\"{i}ve and
J+aB fail to achieve nominal coverage (80\%), rendering them unsuitable for reliable
clinical decision-making. Among the valid methods, Standard CP and CV+ exhibit
comparable performance. Given this parity, we select Standard CP as the primary
baseline. This choice reflects its status as the established standard for split
conformal prediction. By benchmarking against this fundamental method, we can most
clearly isolate the impact of our adaptive scaling mechanism, evaluating its efficiency
gains directly against the standard methods used in practice.

\textbf{The Structural Limitation of Standard Baselines.} By mathematical design,
standard conformal methods assume homoscedasticity, applying a global non-conformity
threshold across the entire feature space. Consequently, they structurally default to
a Cascade Ratio near 1.0, as their interval lengths cannot respond to local epistemic
risk. While this guarantees uniform interval sizes, the CR metric exposes the clinical
danger of this rigidity: these standard models are forced to offer identical safety
buffers to both straightforward and highly ambiguous patient profiles.

While mathematically valid, this results in over-coverage (84.0\% [0.801--0.879]),
revealing that standard baselines achieve safety through rigid, global conservatism
rather than patient-level precision. CASCADE, by dynamically routing uncertainty,
calibrates tightly to the 80.1\% target. The variance in CASCADE's interval length
([0.135--0.162]) is not statistical noise but the intentional result of
uncertainty-driven scaling, providing optimized precision for the patients that
need it most.

\subsection{Mechanism of Adaptivity: Group-wise Analysis}

Global metrics often mask significant performance disparities between easy (low
uncertainty) and difficult (high uncertainty) cases. While strict conditional coverage
is theoretically ideal, recent work suggests it can yield overly conservative intervals
in practice due to data scarcity in high-risk regions \citep{Ding2023}. To explicitly
operationalize the trade-off between routine clinical precision and high-stakes safety,
we adopted a stratified approach, grouping patients into tertiles based on their
Venn-Abers uncertainty scores ($u_{\text{VA}}$). This distinction allows us to decouple
performance for low and high uncertainty patients (bottom 33\% and top 33\% respectively).

\begin{table}[t]
  \caption{Stratified Performance Analysis. Comparisons are made against the Standard
  CP baseline length (0.113). CASCADE dynamically reallocates this budget: it tightens
  intervals for confident cases (Low) by 38.9\% while creating a safety buffer for the
  high-risk patients (Top 33\%).}
  \label{tab:stratified}
  \begin{center}
    \begin{small}
      \begin{tabular}{llccc}
        \toprule
        Uncertainty & Method & Cov. & Length & $\Delta$ Length$^\dagger$ \\
        \midrule
        Low    & Standard CP & 81.1\% & 0.113 & --- \\
               & \textbf{Cascade}     & \textbf{69.7\%} & \textbf{0.069} & \textbf{$-$38.9}\% \\
        \midrule
        Medium & Standard CP & 86.5\% & 0.113 & --- \\
               & \textbf{Cascade}     & \textbf{82.0\%} & \textbf{0.100} & \textbf{$-$10.9\%} \\
        \midrule
        High   & Standard CP & 85.4\% & 0.113 & --- \\
               & \textbf{Cascade}     & \textbf{91.7\%} & \textbf{0.292} & \textbf{$+$158.9\%} \\
        \bottomrule
      \end{tabular}
    \end{small}
  \end{center}
  \vskip -0.1in
  {\footnotesize $^\dagger$Relative to Standard CP Baseline Length (0.113).}
\end{table}

Table~\ref{tab:stratified} contrasts the behavior of Standard CP (fixed length 0.113)
against the adaptive response of CASCADE ($\beta = 0.7$), revealing the core cascade
effect. For \textbf{low-uncertainty} patients, CASCADE identifies the standard length
as unnecessarily conservative and automatically tightens to 0.069 ($-$38.9\%), offering
clinicians more precise guidance for confident adjustments. For \textbf{high-uncertainty}
patients (top 33\%), CASCADE creates a dynamic safety net by expanding to 0.292
(+158.9\%), simultaneously increasing coverage from 85.4\% to 91.7\% and signaling
that ambiguous cases warrant additional caution before adjusting medication.

\subsection{Quantitative Validation of Adaptivity}

To validate the statistical significance of the proposed framework and verify the
drivers of interval adaptivity, we employ the following validation tests:

\textbf{Distributional Divergence (KS Test):} The KS test yielded a statistic of
$D = 0.62$ ($p < 10^{-54}$), rejecting the null hypothesis that CASCADE and the
Standard CP baseline produce equivalent interval distributions. This confirms that
CASCADE fundamentally alters the prediction structure, shifting from a static safety
margin to a dynamic distribution.

\textbf{Driver of Adaptivity (Spearman Correlation):} To verify that this distinct
behavior is driven by the uncertainty scores, we evaluate the correlation between
interval length and the Venn-Abers score ($u_{\text{VA}}$). CASCADE demonstrates a
near-perfect Spearman correlation of $\rho = 0.999$ ($p < 0.001$), validating that
prediction intervals are directly and monotonically responsive to the detected epistemic
risk.

\textbf{Global Utility (Winkler Score):} The adaptivity of CASCADE introduces an
expected trade-off in global efficiency. The framework achieved a Winkler score of
0.392, compared to the baseline's 0.282. This increase in the aggregate penalty
reflects CASCADE's interval length adaptivity to high-risk, uncertain predictions.
While the baseline maintains tighter average intervals by ignoring local uncertainty,
CASCADE prioritizes clinical safety by explicitly inflating bounds where epistemic risk
is high, providing the essential adaptive signaling required for safe downstream clinical
intervention.

A detailed analysis of Mondrian sensitivity across $K \in \{3, 5, 7\}$ stratification
bins, and the corresponding robustness of Continuous CASCADE, is provided in
Appendix~\ref{sec:granularity}.

\subsection{Beta Ablation}
\label{sec:beta}

We perform a comprehensive ablation study on the scaling parameter to characterize
the trade-off between coverage stability and interval length. Figure~\ref{fig:beta}
visualizes this dynamic, while quantitative results are detailed in
Table~\ref{tab:full_beta} in Appendix~\ref{sec:beta_ablation}.

\begin{figure}[t]
  \vskip 0.2in
  \begin{center}
    \centerline{\includegraphics[width=\columnwidth]{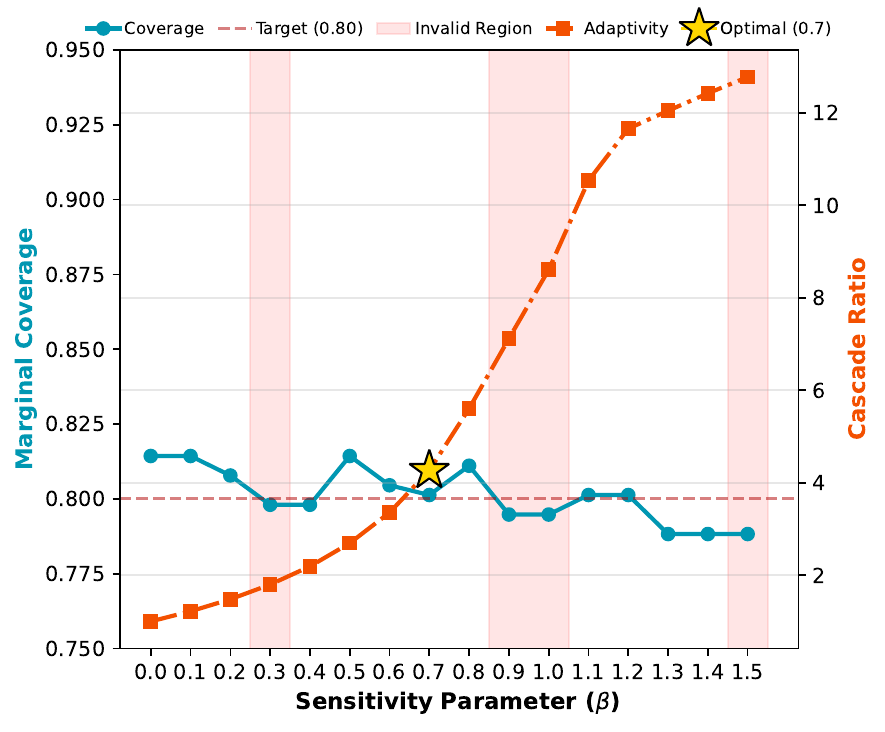}}
    \caption{$\beta$ Ablation Study. The trade-off between safety and adaptivity.
    The dark red line denotes nominal coverage (target 80\%), while the orange line
    denotes the Cascade Ratio (adaptivity). The optimal parameter $\beta = 0.7$ (gold
    star) maximizes adaptivity (CR 4.23) while strictly maintaining adequate coverage
    (80.1\%). Lower values are inefficient while higher values violate coverage.}
    \label{fig:beta}
  \end{center}
  \vskip -0.2in
\end{figure}

Our analysis identifies $\beta = 0.7$ as the optimal value. At this threshold, the
model maximizes adaptivity (Cascade Ratio 4.23) while strictly maintaining valid
marginal coverage (80.1\%).

Lower values ($\beta \leq 0.5$) provide valid coverage but reduced adaptive resolution
(CR $< 3.0$). Conversely, aggressive scaling ($\beta \in [0.9, 1.0]$) violates the
coverage guarantee. While extreme scaling ($\beta = 1.2$) restores validity, it results
in an average interval length of 0.267 (more than double the baseline length) rendering
the predictions inefficient for clinical utility. Thus, $\beta = 0.7$ represents the
maximal adaptivity achievable within the safety constraints of the task.

\section{Discussion}
\label{sec:discussion}

\textbf{Uncertainty Propagation in Two-Stage Systems.} Two-stage clinical frameworks
successfully decouple intervention assessment from dosage prediction, but they typically
treat the subsequent regression task as statistically independent. Our results challenge
this compartmentalization: the epistemic uncertainty generated during the assessment
phase is a vital signal that should not be discarded. CASCADE enables the downstream
predictor to dynamically adapt its intervals based on upstream ambiguity, ensuring that
the final recommendation reflects the cumulative uncertainty of the entire pipeline.

\textbf{The Stability/Adaptivity Tradeoff.} A common critique of adaptive conformal
methods is the loss of coverage stability across subgroups. While Standard CP achieves
uniform coverage (84\%) across all risk profiles, our analysis suggests this
``stability'' results in unnecessarily wide intervals. As \citet{Ding2023} warns,
strict conditional coverage can lead to overly conservative predictions in data-scarce
regions. Instead, CASCADE adopts local adaptivity. By allowing the interval length to
fluctuate, we accept a marginal coverage adjustment in the low-risk majority (69.7\%)
to secure a significant reduction in interval length (38.9\%). While this targeted
local adaptivity inherently results in a higher global Winkler score compared to
baseline, this trade-off prioritizes high-stakes clinical utility over theoretical
uniformity.

\textbf{Limits of Discrete Stratification.} Our analysis reveals a fundamental tension
between granularity and statistical power in discrete frameworks. While Mondrian CP
offers intuitive group-conditional guarantees, it suffers from sample size
fragmentation. As shown in Appendix~\ref{sec:granularity}, increasing the bin count
to $K = 7$ achieved a higher Cascade Ratio (4.63) but caused Mondrian's average
interval length to inflate to 0.170 (a 44\% increase compared to 0.118 at $K = 3$)
due to reduced effective sample size per bin ($N_{\text{cal}}/K$). In contrast,
Continuous CASCADE decouples uncertainty scaling from quantile estimation, maintaining
strong adaptivity (CR 6.83 at $K = 7$) without the fragmentation penalty.

\textbf{Clinical Implications.} The lack of adaptivity in Standard CP (CR 1.00) poses
a latent safety risk by presenting identical interval lengths for both clear-cut and
ambiguous cases. This false sense of certainty can lead to dangerous over-reliance on
model outputs and unfortunate patient outcomes. The CASCADE framework addresses this
by acting as an automated safeguard: for high-uncertainty patients, it expands the
prediction interval and explicitly signals to the neurologist that the case warrants
caution, prompting investigation of confounding factors (e.g., protein intake or sleep
quality) before adjusting medication. This aligns model behavior with the principle of
\textit{primum non nocere} (first, do no harm). Crucially, CASCADE does not achieve
this safety by inflating intervals globally. Instead, it employs a targeted mechanism,
utilizing optimized prediction intervals to preserve high precision for the majority of
confident cases while reserving adaptive safety buffers for highly ambiguous cases.

\textbf{Beyond Parkinson's Disease.} While validated on PD medication adjustment, CASCADE is fundamentally domain-agnostic. It addresses the reliability gap inherent to any two-stage pipeline, where a discrete intervention assessment precedes a continuous quantification task. Examples include Deep Brain Stimulation, where discrete selection of the surgical target (globus pallidus interna vs. subthalamic nucleus) precedes fine-tuning of voltage and frequency \citep{Ramirez-Zamora2018}. Similarly, in dystonia, identifying affected muscle groups is a prerequisite for calculating botulinum toxin dosage \citep{Dressler2021}. By propagating the epistemic uncertainty from the initial assessment to the downstream prediction, CASCADE offers a scalable mechanism for building trustworthy, uncertainty-aware AI systems across high-stakes medical domains.

\section{Limitations and Future Work}
\label{sec:limitations}

Our framework employs symmetric scaling, reflecting the clinical reality that
both under and over-medication are detrimental. We aim to explore asymmetric scaling
for profiles where one risk strictly outweighs the other, and to add a formal
abstention mechanism enabling the model to defer highly ambiguous cases to human
experts. Finally, we aim to validate CASCADE across the PD continuum, ensuring
robustness against the longitudinal physiological shifts inherent to disease
progression.

\section{Conclusion}
\label{sec:conclusion}

We present CASCADE, a novel uncertainty quantification framework that resolves the
tension between classification precision and regression reliability. By leveraging
Venn-Abers uncertainty to continuously scale prediction intervals, CASCADE helps
overcome the ``one-size-fits-all'' nature of PD medication management. By dynamically
calibrating the prediction interval to the upstream confidence signal, CASCADE achieves
a four-fold improvement in adaptivity (CR 4.23 vs 1.00) while maintaining
coverage guarantees. This ensures that AI-driven dosage recommendations
are not only accurate, but contextually aware: providing optimized prediction intervals
for straightforward patients and a necessary adaptive safety buffer for complex ones.
Ultimately, CASCADE demonstrates that adaptivity is not just a technical enhancement
but a clinical necessity.

\section*{Impact Statement}

CASCADE is designed to surface uncertainty rather than suppress it, ensuring clinicians
receive explicit warning signals before acting on ambiguous predictions. As with all
clinical AI, prospective validation and regulatory review are necessary before
deployment. There are many potential societal consequences of our work, none which we
feel must be specifically highlighted beyond this note.

\bibliography{example_paper}

@article{Svensson2018,
   author = {Fredrik Svensson and Natalia Aniceto and Ulf Norinder and Isidro Cortes-Ciriano and Ola Spjuth and Lars Carlsson and Andreas Bender},
   doi = {10.1021/acs.jcim.8b00054},
   issn = {1549960X},
   issue = {5},
   journal = {Journal of Chemical Information and Modeling},
   month = {5},
   pages = {1132-1140},
   pmid = {29701973},
   publisher = {American Chemical Society},
   title = {Conformal Regression for Quantitative Structure-Activity Relationship Modeling - Quantifying Prediction Uncertainty},
   volume = {58},
   year = {2018}
}

@article{Papadopoulos2017,
   author = {Harris Papadopoulos and Efthyvoulos Kyriacou and Andrew Nicolaides},
   doi = {10.1007/s00521-016-2590-3},
   issn = {09410643},
   issue = {6},
   journal = {Neural Computing and Applications},
   month = {6},
   pages = {1209-1223},
   publisher = {Springer London},
   title = {Unbiased confidence measures for stroke risk estimation based on ultrasound carotid image analysis},
   volume = {28},
   year = {2017}
}

@article{Kim2020,
   author = {Byol Kim and Chen Xu and Rina Barber},
   journal = {Advances in Neural Information Processing Systems},
   pages = {4138-4149},
   title = {Predictive inference is free with the jackknife+-after-bootstrap},
   volume = {33},
   year = {2020}
}

@article{Barber2021,
   author = {Rina Foygel Barber and Emmanuel J Candes and Aaditya Ramdas and Ryan J Tibshirani},
   issue = {1},
   journal = {The Annals of Statistics},
   pages = {486-507},
   publisher = {JSTOR},
   title = {Predictive inference with the jackknife+},
   volume = {49},
   year = {2021}
}

@article{Watts2021,
   author = {Jeremy Watts and Anahita Khojandi and Rama Vasudevan and Fatta B. Nahab and Ritesh A. Ramdhani},
   doi = {10.3390/s21103553},
   issn = {14248220},
   issue = {10},
   journal = {Sensors},
   month = {5},
   pmid = {34065245},
   publisher = {MDPI AG},
   title = {Improving medication regimen recommendation for parkinson’s disease using sensor technology},
   volume = {21},
   year = {2021}
}

@article{Shamir2015,
   author = {Reuben R. Shamir and Trygve Dolber and Angela M. Noecker and Benjamin L. Walter and Cameron C. McIntyre},
   doi = {10.1016/j.brs.2015.06.003},
   issn = {18764754},
   issue = {6},
   journal = {Brain Stimulation},
   month = {11},
   pages = {1025-1032},
   pmid = {26140956},
   publisher = {Elsevier Inc.},
   title = {Machine learning approach to optimizing combined stimulation and medication therapies for Parkinson's disease},
   volume = {8},
   year = {2015}
}

@article{Salmanpour2023,
   author = {Mohammad R. Salmanpour and Mahdi Hosseinzadeh and Mahya Bakhtiyari and Mehdi Maghsudi and Arman Rahmim},
   doi = {10.1002/ima.22868},
   issn = {10981098},
   issue = {4},
   journal = {International Journal of Imaging Systems and Technology},
   month = {7},
   pages = {1437-1449},
   publisher = {John Wiley and Sons Inc},
   title = {Prediction of drug amount in Parkinson's disease using hybrid machine learning systems and radiomics features},
   volume = {33},
   year = {2023}
}

@article{Kim2021,
   author = {Yejin Kim and Jessika Suescun and Mya C. Schiess and Xiaoqian Jiang},
   doi = {10.1038/s41598-021-88619-4},
   issn = {20452322},
   issue = {1},
   journal = {Scientific Reports},
   month = {12},
   pmid = {33927277},
   publisher = {Nature Research},
   title = {Computational medication regimen for Parkinson’s disease using reinforcement learning},
   volume = {11},
   year = {2021}
}

@article{Olanow2006,
   author = {C Warren Olanow and Jose A Obeso and Fabrizio Stocchi},
   issue = {8},
   journal = {The Lancet Neurology},
   pages = {677-687},
   publisher = {Elsevier},
   title = {Continuous dopamine-receptor treatment of Parkinson's disease: scientific rationale and clinical implications},
   volume = {5},
   year = {2006}
}

@misc{Jenner2008,
   author = {Peter Jenner},
   doi = {10.1038/nrn2471},
   issn = {1471003X},
   issue = {9},
   journal = {Nature Reviews Neuroscience},
   month = {9},
   pages = {665-677},
   pmid = {18714325},
   title = {Molecular mechanisms of L-DOPA-induced dyskinesia},
   volume = {9},
   year = {2008}
}

@book{vovk2005,
   author = {Vladimir Vovk and Alexander Gammerman and Glenn Shafer},
   publisher = {Springer},
   title = {Algorithmic Learning in a Random World},
   year = {2005}
}

@article{Tomlinson2010,
   author = {Claire L. Tomlinson and Rebecca Stowe and Smitaa Patel and Caroline Rick and Richard Gray and Carl E. Clarke},
   doi = {10.1002/mds.23429},
   issn = {08853185},
   issue = {15},
   journal = {Movement Disorders},
   month = {11},
   pages = {2649-2653},
   pmid = {21069833},
   title = {Systematic review of levodopa dose equivalency reporting in Parkinson's disease},
   volume = {25},
   year = {2010}
}

@inproceedings{Ying2019,
   author = {Xue Ying},
   doi = {10.1088/1742-6596/1168/2/022022},
   issn = {17426596},
   issue = {2},
   booktitle = {Journal of Physics: Conference Series},
   month = {3},
   publisher = {Institute of Physics Publishing},
   title = {An Overview of Overfitting and its Solutions},
   volume = {1168},
   year = {2019}
}

@article{Wu2019,
   author = {Jia Wu and Xiu Yun Chen and Hao Zhang and Li Dong Xiong and Hang Lei and Si Hao Deng},
   doi = {10.11989/JEST.1674-862X.80904120},
   issn = {2666223X},
   issue = {1},
   journal = {Journal of Electronic Science and Technology},
   month = {3},
   pages = {26-40},
   publisher = {Univ. of Electronic Science and Technology of China},
   title = {Hyperparameter optimization for machine learning models based on Bayesian optimization},
   volume = {17},
   year = {2019}
}

@article{Espay2016,
   author = {Alberto J. Espay and Paolo Bonato and Fatta B. Nahab and Walter Maetzler and John M. Dean and Jochen Klucken and Bjoern M. Eskofier and Aristide Merola and Fay Horak and Anthony E. Lang and Ralf Reilmann and Joe Giuffrida and Alice Nieuwboer and Malcolm Horne and Max A. Little and Irene Litvan and Tanya Simuni and E. Ray Dorsey and Michelle A. Burack and Ken Kubota and Anita Kamondi and Catarina Godinho and Jean Francois Daneault and Georgia Mitsi and Lothar Krinke and Jeffery M. Hausdorff and Bastiaan R. Bloem and Spyros Papapetropoulos},
   doi = {10.1002/mds.26642},
   issn = {15318257},
   issue = {9},
   journal = {Movement Disorders},
   month = {9},
   pages = {1272-1282},
   pmid = {27125836},
   publisher = {John Wiley and Sons Inc.},
   title = {Technology in Parkinson's disease: Challenges and opportunities},
   volume = {31},
   year = {2016}
}

@techReport{Shafer2008,
   author = {Glenn Shafer and Vladimir Vovk},
   journal = {Journal of Machine Learning Research},
   pages = {371-421},
   title = {A Tutorial on Conformal Prediction},
   volume = {9},
   year = {2008}
}

@article{Angelopoulos2023,
   author = {Anastasios N. Angelopoulos and Stephen Bates},
   doi = {10.1561/2200000101},
   issn = {1935-8237},
   issue = {4},
   journal = {Foundations and Trends® in Machine Learning},
   pages = {494-591},
   publisher = {Now Publishers},
   title = {Conformal Prediction: A Gentle Introduction},
   volume = {16},
   year = {2023}
}

@article{Angelopoulos2021,
   author = {Anastasios N. Angelopoulos and Stephen Bates},
   journal = {arXiv preprint},
   month = {7},
   title = {A Gentle Introduction to Conformal Prediction and Distribution-Free Uncertainty Quantification},
   url = {http://arxiv.org/abs/2107.07511},
   year = {2021}
}

@article{Beli2019,
   author = {M. Belić and Vladislava Bobić and Milica Badža and Nikola Šolaja and Milica Đurić-Jovičić and Vladimir S. Kostić},
   doi = {10.1016/j.clineuro.2019.105442},
   issn = {18726968},
   journal = {Clinical Neurology and Neurosurgery},
   month = {9},
   pmid = {31351213},
   publisher = {Elsevier B.V.},
   title = {Artificial intelligence for assisting diagnostics and assessment of Parkinson's disease—A review},
   volume = {184},
   year = {2019}
}

@techReport{Ng2004,
   author = {Andrew Y Ng},
   title = {Feature selection, L 1 vs. L 2 regularization, and rotational invariance},
   year = {2004}
}

@inproceedings{Chen2016,
   author = {Tianqi Chen and Carlos Guestrin},
   doi = {10.1145/2939672.2939785},
   isbn = {9781450342322},
   booktitle = {Proceedings of the ACM SIGKDD International Conference on Knowledge Discovery and Data Mining},
   month = {8},
   pages = {785-794},
   publisher = {Association for Computing Machinery},
   title = {XGBoost: A scalable tree boosting system},
   volume = {13-17-August-2016},
   year = {2016}
}

@article{Armstrong2020,
   author = {Melissa J. Armstrong and Michael S. Okun},
   doi = {10.1001/jama.2019.22360},
   issn = {15383598},
   issue = {6},
   journal = {JAMA - Journal of the American Medical Association},
   month = {2},
   pages = {548-560},
   pmid = {32044947},
   publisher = {American Medical Association},
   title = {Diagnosis and Treatment of Parkinson Disease: A Review},
   volume = {323},
   year = {2020}
}

@article{Vovk2012,
  title={Venn-abers predictors},
  author={Vovk, Vladimir and Petej, Ivan},
  journal={arXiv preprint arXiv:1211.0025},
  year={2012}
}

@InProceedings{diaz-rincon25a,
  title = 	 {Uncertainty-Aware Prediction of {Parkinson's} Disease Medication Needs: A Two-Stage Conformal Prediction Approach},
  author =       {Diaz-Rincon, Ricardo and Liang, Muxuan and Ramirez-Zamora, Adolfo and Shickel, Benjamin},
  booktitle = 	 {Proceedings of the 10th Machine Learning for Healthcare Conference},
  year = 	 {2025},
  volume = 	 {298},
  series = 	 {Proceedings of Machine Learning Research},
  month = 	 {15--16 Aug},
  publisher =    {PMLR},
  url = {https://proceedings.mlr.press/v298/diaz-rincon25a.html}
}

@inproceedings{diaz-rincon24,
  title = {{AI} for Personalized Medication Management in {Parkinson's} Disease ({PD})},
  author={Diaz-Rincon, R and Khoshbouei, H and Shickel, B},
  booktitle={MOVEMENT DISORDERS},
  volume={39},
  pages={S48--S49},
  year={2024},
  organization={WILEY 111 RIVER ST, HOBOKEN 07030-5774, NJ USA}
}

@inproceedings{diaz-rincon25b,
  title   = {Improved Decision-Making for In-Hospital Medication Management in {Parkinson's} Disease},
  author={Diaz-Rincon, R and Liang, M and Ramirez-Zamora, A and Shickel, B},
  booktitle={MOVEMENT DISORDERS},
  volume={40},
  pages={S23--S24},
  year={2025},
  organization={WILEY 111 RIVER ST, HOBOKEN 07030-5774, NJ USA}
}

@article{Lei2018,
  title={Distribution-free predictive inference for regression},
  author={Lei, Jing and G’Sell, Max and Rinaldo, Alessandro and Tibshirani, Ryan J and Wasserman, Larry},
  journal={Journal of the American Statistical Association},
  volume={113},
  number={523},
  pages={1094--1111},
  year={2018},
  publisher={Taylor \& Francis}
}

@article{newman2009parkinsonism,
  title={The parkinsonism-hyperpyrexia syndrome},
  author={Newman, Edward J and Grosset, Donald G and Kennedy, Peter GE},
  journal={Neurocritical Care},
  volume={10},
  number={1},
  pages={136--140},
  year={2009},
  publisher={Springer}
}

@article{taylor2016neuropsychiatric,
  title={Neuropsychiatric complications of Parkinson disease treatments: importance of multidisciplinary care},
  author={Taylor, Jacob and Anderson, William S and Brandt, Jason and Mari, Zoltan and Pontone, Gregory M},
  journal={The American Journal of Geriatric Psychiatry},
  volume={24},
  number={12},
  pages={1171--1180},
  year={2016},
  publisher={Elsevier}
}

@article{Papadopoulos2011,
  title={Regression conformal prediction with nearest neighbours},
  author={Papadopoulos, Harris and Vovk, Vladimir and Gammerman, Alex},
  journal={Journal of Artificial Intelligence Research},
  volume={40},
  pages={815--840},
  year={2011}
}

@inproceedings{Guo2017,
  title={On calibration of modern neural networks},
  author={Guo, Chuan and Pleiss, Geoff and Sun, Yu and Weinberger, Kilian Q},
  booktitle={International conference on machine learning},
  pages={1321--1330},
  year={2017},
  organization={PMLR}
}

@article{Ramirez-Zamora2018,
  title={Globus pallidus interna or subthalamic nucleus deep brain stimulation for Parkinson disease: a review},
  author={Ramirez-Zamora, Adolfo and Ostrem, Jill L},
  journal={JAMA neurology},
  volume={75},
  number={3},
  pages={367--372},
  year={2018},
  publisher={American Medical Association}
}

@article{Dressler2021,
  title={Consensus guidelines for botulinum toxin therapy: general algorithms and dosing tables for dystonia and spasticity},
  author={Dressler, Dirk and Altavista, Maria Concetta and Altenmueller, Eckart and Bhidayasiri, Roongroj and Bohlega, Saeed and Chana, Pedro and Chung, Tae Mo and Colosimo, Carlo and Fheodoroff, Klemens and Garcia-Ruiz, Pedro J and others},
  journal={Journal of Neural Transmission},
  volume={128},
  number={3},
  pages={321--335},
  year={2021},
  publisher={Springer}
}

@article{Ding2023,
  title={Class-conditional conformal prediction with many classes},
  author={Ding, Tiffany and Angelopoulos, Anastasios and Bates, Stephen and Jordan, Michael and Tibshirani, Ryan J},
  journal={Advances in neural information processing systems},
  volume={36},
  pages={64555--64576},
  year={2023}
}
\bibliographystyle{icml2026}

\newpage
\appendix
\onecolumn

\section{Patient Cohort and Medication Details}
\label{sec:cohort}

\subsection{Demographics}

Table~\ref{tab:demographics} details the demographic characteristics of the study
cohort ($N = 631$). The population is predominantly elderly (Mean Age 78.0) and male
(64.5\%), consistent with Parkinson's Disease epidemiology.

\begin{table}[h]
  \caption{Demographic Characteristics of the Study Cohort.}
  \label{tab:demographics}
  \begin{center}
    \begin{tabular}{llcc}
      \toprule
      Characteristic & Category & Number & Percentage (\%) \\
      \midrule
      Total Patients & & 631 & 100.00 \\
      Mean Age, Years (SD) & & 77.98 (10.59) & \\
      Gender & Male & 407 & 64.50 \\
             & Female & 224 & 35.50 \\
      Race   & White & 526 & 83.62 \\
             & Black & 63  & 10.01 \\
             & Other & 34  & 5.41  \\
             & Asian & 5   & 0.79  \\
             & Multiracial & 1 & 0.16 \\
      Ethnicity & Not Hispanic & 598 & 95.07 \\
                & Hispanic & 31 & 4.93 \\
      \bottomrule
    \end{tabular}
  \end{center}
\end{table}

\subsection{Medications}

Table~\ref{tab:medications} lists the antiparkinsonian medications administered in our
patient cohort.

\begin{table}[h]
  \caption{Classification of antiparkinsonian medications administered during the study.}
  \label{tab:medications}
  \begin{center}
    \begin{tabular}{ll}
      \toprule
      Drug Class & Medications \\
      \midrule
      Levodopa Formulations & Carbidopa-Levodopa, Carbidopa-Levodopa ER, \\
                            & Carbidopa-Levodopa-Entacapone, Duopa \\
      Dopamine Agonists & Pramipexole, Pramipexole ER, Ropinirole, \\
                        & Rotigotine, Apomorphine, Bromocriptine, Cabergoline \\
      MAO-B Inhibitors  & Rasagiline, Selegiline, Safinamide \\
      COMT Inhibitors   & Entacapone, Tolcapone \\
      Anticholinergics  & Benztropine, Trihexyphenidyl \\
      NMDA Antagonists  & Amantadine, Amantadine ER \\
      Adenosine Antagonists & Istradefylline \\
      \bottomrule
    \end{tabular}
  \end{center}
\end{table}

\section{Model Implementation and Conformal Framework}
\label{sec:model}

\subsection{Underlying Predictors (XGBoost)}

To ensure that the uncertainty quantification results reflect the framework's
performance rather than model instability, we utilized strong underlying predictors.
For both classification and regression, we employed XGBoost models \citep{Chen2016}
with hyperparameters optimized via GridSearchCV and Bayesian Optimization \citep{Wu2019}.

Key hyperparameters included a learning rate of 0.1, max depth of 7, and 700 boosting
rounds. We applied L1/L2 regularization \citep{Ng2004} (alpha: 0.1, lambda: 1) and
early stopping to prevent overfitting \citep{Ying2019}. This configuration demonstrated
strong predictive capabilities, achieving an AUC of 0.96 for the classification task
and high precision for the dosage prediction ($R^2 = 0.9801$, RMSE$=0.0995$,
MAE$=0.0380$).

\subsection{Conformal Prediction Baselines}
\label{sec:cp_framework}

We compared the proposed method against four standard benchmarks.

\textbf{Standard Conformal Prediction:} We employ the Standard Conformal Prediction
framework. The available training data is split into a proper training set
$\mathcal{D}_{\text{train}}$ and a calibration set $\mathcal{D}_{\text{cal}}$. The
model is fitted on $\mathcal{D}_{\text{train}}$, and non-conformity scores
$s_i = |y_i - \hat{\mu}(x_i)|$ are computed on $\mathcal{D}_{\text{cal}}$. The
prediction interval is constructed as $\hat{\mu}(x_{\text{new}}) \pm \hat{q}_{1-\alpha}$,
where $\hat{q}_{1-\alpha}$ is the $(1-\alpha)$ quantile of the calibration scores.

\textbf{Na\"{i}ve:} This method constructs fixed-length intervals based solely on the
empirical distribution of residuals. The prediction interval is defined as
$\hat{\mu}(x_{\text{new}}) \pm q_{1-\alpha}$, where $q_{1-\alpha}$ is the $(1-\alpha)$
quantile of the absolute errors on the calibration set. Unlike conformal prediction,
the Na\"{i}ve approach lacks finite-sample coverage guarantees and does not account for
the statistical variability introduced by the calibration step.

\textbf{Cross Validation Conformal (CV+):} This method employs $K$-fold
cross-validation ($K = 10$) to leverage multiple calibration sets. While CV+ provides
a theoretical coverage guarantee of $1 - 2\alpha$, it is computationally expensive
($K$ model fits) and often produces conservative intervals \citep{Barber2021}.

\textbf{Jackknife+ After Bootstrap (J+aB):} Designed specifically for ensemble methods.
It reuses the bootstrap samples from the ensemble training to create leave-one-out
predictions. J+aB provides intervals ``for free'' with ensemble methods but may exhibit
instability in small datasets \citep{Kim2020}.

\section{Extended Mondrian Results}
\label{sec:mondrian_results}

We provide detailed performance metrics for the discrete stratification approach
(Mondrian Conformal Prediction) discussed in the main text. Table~\ref{tab:mondrian_k3}
presents the coverage, interval length, and adaptivity signal for Mondrian ($K = 3$).

\begin{table}[h]
  \caption{Performance Metrics for Mondrian Conformal Prediction ($K = 3$). The method
  stratifies patients into three discrete bins based on uncertainty. While coverage is
  valid, the adaptivity signal is coarse compared to the continuous approach.}
  \label{tab:mondrian_k3}
  \begin{center}
    \begin{tabular}{lccc}
      \toprule
      Uncertainty & Coverage & Length & Adaptivity \\
      \midrule
      Low    & 80.3\% & 0.090 & Moderate \\
      Medium & 88.8\% & 0.090 & Moderate \\
      High   & 92.7\% & 0.181 & High \\
      \midrule
      \multicolumn{3}{l}{Overall Cascade Ratio (CR)} & 2.02 \\
      \bottomrule
    \end{tabular}
  \end{center}
\end{table}

\section{Impact of Stratification Granularity}
\label{sec:granularity}

To ensure our conclusions are not artifacts of the chosen discretization level ($K=3$),
we evaluate the impact of increasing the number of uncertainty strata ($K \in \{5, 7\}$).
Table~\ref{tab:mondrian_sensitivity} presents the performance of Mondrian as the number
of bins increases. Moving from $K=3$ to $K=5$, the model maintains a stable Cascade
Ratio (CR) while holding coverage relatively close to the target guarantee. However, as
$K$ is pushed to 7, the framework exhibits a sharp spike in adaptivity (CR 4.63). This
demonstrates the framework's underlying capacity to scale intervals for highly uncertain
cohorts when given higher resolution.

\begin{table}[h]
  \caption{Mondrian Sensitivity. As the number of strata ($K$) increases, the framework
  demonstrates an increased capacity for adaptivity, peaking at a CR of 4.63 at $K=7$.}
  \label{tab:mondrian_sensitivity}
  \begin{center}
    \begin{tabular}{lccc}
      \toprule
      Configuration & Cov.\ (\%) & Length & CR \\
      \midrule
      Mondrian ($K=3$) & 86.6\% & 0.118 & 2.02 \\
      Mondrian ($K=5$) & 84.0\% & 0.121 & 1.98 \\
      Mondrian ($K=7$) & 85.3\% & 0.170 & 4.63 \\
      \bottomrule
    \end{tabular}
  \end{center}
\end{table}

\subsection{Continuous CASCADE (Evaluation Robustness)}

To verify that our efficiency gains are not sensitive to the choice of evaluation bins,
we analyze the Cascade Ratio across varying granularities (from tertiles to septiles).
As shown in Table~\ref{tab:cascade_robustness}, the performance remains stable across
all resolutions. While we report results for $K = 3$ in the main text to align with
standard clinical risk strata (Low, Medium, High), CASCADE maintains high adaptivity
even at finer resolutions. Notably, even at the finest granularity ($K = 7$), the
model maintains a Cascade Ratio of 6.83, confirming that the uncertainty scaling is a
fundamental property of the signal rather than an artifact of discretization.

\begin{table}[h]
  \caption{Continuous CASCADE Robustness. The adaptivity signal remains strong across
  different evaluation resolutions, consistently outperforming the discrete baseline.}
  \label{tab:cascade_robustness}
  \begin{center}
    \begin{tabular}{lcc}
      \toprule
      Evaluation Split & Comparison & CR \\
      \midrule
      Tertiles ($K=3$)  & Top 33\% vs Bottom 33\% & 4.23 \\
      Quintiles ($K=5$) & Top 20\% vs Bottom 20\% & 5.85 \\
      Septiles ($K=7$)  & Top 14\% vs Bottom 14\% & 6.83 \\
      \bottomrule
    \end{tabular}
  \end{center}
\end{table}

\section{Extended Beta Ablation}
\label{sec:beta_ablation}

To validate the stability of the CASCADE framework, we performed a fine-grained
ablation study on the scaling parameter $\beta$. The range $\beta \in [0.0, 1.5]$ was
chosen to encompass the full theoretical spectrum: from $\beta = 0$ (homoscedastic
assumption, effectively Standard CP) through $\beta = 1$ (linear scaling proportional
to uncertainty), up to $\beta = 1.5$ (aggressive supra-linear penalization).
Table~\ref{tab:full_beta} shows these results.

\begin{table}[h]
  \caption{Full $\beta$ Ablation. We vary the scaling factor $\beta \in [0.0, 1.5]$.
  The selected value $\beta = 0.7$ (bold) is chosen by the Cascade Ratio (CR), which
  quantifies the adaptivity gain relative to the base interval length. $\beta = 0.7$
  achieves the optimal CR (4.23) while maintaining valid marginal coverage. Values of
  $\beta \geq 0.9$ result in coverage collapse for the low-uncertainty cohort.}
  \label{tab:full_beta}
  \begin{center}
    \begin{small}
      \begin{tabular}{lcccccccc}
        \toprule
        & \multicolumn{2}{c}{Low Unc.} & \multicolumn{2}{c}{Med.\ Unc.}
        & \multicolumn{2}{c}{High Unc.} & & \\
        \cmidrule(lr){2-3}\cmidrule(lr){4-5}\cmidrule(lr){6-7}
        $\beta$ & Cov. & Len. & Cov. & Len. & Cov. & Len. & CR & Status \\
        \midrule
        0.0 & 80.3\% & 0.128 & 83.1\% & 0.128 & 81.2\% & 0.128 & 1.00 & No Adaptivity \\
        0.1 & 77.9\% & 0.112 & 83.1\% & 0.116 & 84.4\% & 0.137 & 1.22 & Minimal \\
        0.2 & 77.0\% & 0.099 & 82.0\% & 0.106 & 84.4\% & 0.147 & 1.49 & Mild \\
        0.3 & 74.6\% & 0.087 & 82.0\% & 0.097 & 84.4\% & 0.158 & 1.81 & Conservative \\
        0.4 & 73.0\% & 0.086 & 82.0\% & 0.100 & 86.5\% & 0.189 & 2.21 & Conservative \\
        0.5 & 73.8\% & 0.084 & 82.0\% & 0.105 & 90.6\% & 0.228 & 2.71 & Valid \\
        0.6 & 71.3\% & 0.077 & 82.0\% & 0.103 & 90.6\% & 0.258 & 3.36 & Valid \\
        \textbf{0.7} & \textbf{69.7\%} & \textbf{0.069} & \textbf{82.0\%} & \textbf{0.100}
          & \textbf{91.7\%} & \textbf{0.292} & \textbf{4.23} & \textbf{Optimal} \\
        0.8 & 70.5\% & 0.067 & 82.0\% & 0.110 & 93.8\% & 0.369 & 5.49 & Borderline \\
        0.9 & 65.6\% & 0.060 & 83.1\% & 0.108 & 93.8\% & 0.416 & 6.93 & Unsafe \\
        1.0 & 63.1\% & 0.056 & 83.1\% & 0.105 & 96.9\% & 0.469 & 8.44 & Unsafe \\
        1.1 & 63.9\% & 0.057 & 83.1\% & 0.115 & 97.9\% & 0.591 & 10.45 & Unsafe \\
        1.2 & 63.9\% & 0.056 & 83.1\% & 0.113 & 97.9\% & 0.679 & 12.01 & Unsafe \\
        1.3 & 63.9\% & 0.058 & 78.7\% & 0.106 & 97.9\% & 0.733 & 12.56 & Unsafe \\
        1.4 & 63.9\% & 0.058 & 78.7\% & 0.098 & 97.9\% & 0.757 & 12.96 & Unsafe \\
        1.5 & 63.9\% & 0.058 & 78.7\% & 0.092 & 97.9\% & 0.779 & 13.34 & Unsafe \\
        \bottomrule
      \end{tabular}
    \end{small}
  \end{center}
\end{table}

\section{Model Performance on Predicted Patients}
\label{sec:predicted_patients}

To demonstrate the robustness of our approach, we perform additional tests for patients
where $\hat{y} \neq 0$, meaning patients where the classifier predicts a change in
medications. Optimal threshold was calculated using Youden's J. Results are summarized
in Table~\ref{tab:predicted_patients}.

\begin{table}[h]
  \caption{While evaluating on predicted patients, Continuous CASCADE achieves perfect
  marginal coverage while delivering the highest Cascade Ratio (1.43) compared to
  Baseline and Mondrian.}
  \label{tab:predicted_patients}
  \begin{center}
    \begin{tabular}{lcccc}
      \toprule
      Method & Cov. & Len. & CR & $\beta$ \\
      \midrule
      Baseline       & 79.4\% & 0.662  & 1.00 & --- \\
      Mondrian       & 82.1\% & 0.625  & 0.83 & --- \\
      Cont.\ CASCADE & 80\%   & 0.6836 & 1.43 & 0.2 \\
      \bottomrule
    \end{tabular}
  \end{center}
\end{table}

\section{Alternative Underlying Models}
\label{sec:alt_models}

We perform additional experiments using Logistic Regression (Stage 1) and Linear
Regression for Stage 2 ($R^2 = 0.08$). As expected, the high underlying model error
caused global interval lengths to expand. While Mondrian's binning mechanism failed to
calibrate under this noise (72.3\% coverage), CASCADE maintained strict marginal
validity (80.1\% coverage) at the cost of longer intervals. Furthermore, CASCADE
successfully achieved a higher CR (2.56), compared to baseline.

\begin{table}[h]
  \caption{Although with expanded intervals due to model errors, CASCADE achieves the
  highest CR (2.56).}
  \label{tab:alt_models}
  \begin{center}
    \begin{tabular}{lcccc}
      \toprule
      Method & Cov. & Len. & CR & $\beta$ \\
      \midrule
      Baseline       & 78.5\%  & 1.649 & 1.00 & --- \\
      Mondrian       & 72.3\%  & 1.721 & 1.07 & --- \\
      Cont.\ CASCADE & 80.13\% & 2.267 & 2.56 & 0.4 \\
      \bottomrule
    \end{tabular}
  \end{center}
\end{table}

\section{Alpha Ablation}
\label{sec:alpha_ablation}

We perform alpha ablation ($\alpha \in \{0.3, 0.1, 0.05\}$) to choose the most
appropriate value for clinical utility. Cont.\ CASCADE consistently satisfies marginal
coverage across all thresholds (e.g., 90.9\% coverage at $\alpha = 0.1$; 95.4\% at
$\alpha = 0.05$), maintaining tighter calibration and narrower intervals than Mondrian
and baselines. We focus on $\alpha = 0.2$ as it provides optimal clinical utility in
PD medication management. While ($\alpha = 0.05$) preserves adaptivity (CR = 4.23),
its intervals become too wide to guide precise LEDD titration.

\begin{table}[h]
  \caption{$\alpha$ Ablation.}
  \label{tab:alpha_ablation}
  \begin{center}
    \begin{tabular}{llcccc}
      \toprule
      $\alpha$ & Method & Cov & Len & CR & $\beta$ \\
      \midrule
      0.30 & Baseline         & 70.4\% & 0.060 & 1.00 & --- \\
           & Mondrian         & 75.9\% & 0.058 & 1.39 & --- \\
           & Cont.\ CASCADE   & 70.0\% & 0.068 & 1.81 & 0.3 \\
      \midrule
      0.20 & Baseline         & 84.0\% & 0.112 & 1.00 & --- \\
           & Mondrian         & 86.6\% & 0.118 & 2.02 & --- \\
           & Cont.\ CASCADE   & 80.1\% & 0.148 & 4.23 & 0.7 \\
      \midrule
      0.10 & Baseline         & 93.5\% & 0.258 & 1.00 & --- \\
           & Mondrian         & 96.1\% & 0.376 & 2.47 & --- \\
           & Cont.\ CASCADE   & 90.9\% & 0.260 & 2.71 & 0.5 \\
      \midrule
      0.05 & Baseline         & 97.7\% & 0.492 & 1.00 & --- \\
           & Mondrian         & 98.0\% & 0.922 & 2.47 & --- \\
           & Cont.\ CASCADE   & 95.4\% & 0.444 & 4.23 & 0.7 \\
      \bottomrule
    \end{tabular}
  \end{center}
\end{table}

\end{document}